\documentclass[letterpaper, 10 pt, conference]{ieeeconf}
\IEEEoverridecommandlockouts

\usepackage{cite}
\usepackage{amsmath,amssymb,amsfonts}
\usepackage{graphicx}
\usepackage{textcomp}
\usepackage[table]{xcolor}
\usepackage{placeins}
\usepackage{booktabs}
\usepackage{multirow}
\usepackage{tikz}
\usetikzlibrary{arrows.meta,positioning,fit,calc}
\definecolor{SWReference}{HTML}{E7EDF3}
\definecolor{SWRegression}{HTML}{8B1E2D}
\definecolor{SWRegressionTint}{HTML}{F5DDE1}
\definecolor{SWSuperIdeal}{HTML}{145A86}
\definecolor{SWSuperIdealTint}{HTML}{DCECF7}
\definecolor{SWGain}{HTML}{145A86}
\definecolor{SWGainTint}{HTML}{DCECF7}
\definecolor{SWLoss}{HTML}{8B1E2D}
\definecolor{SWLossTint}{HTML}{F5DDE1}

\def\BibTeX{{\rm B\kern-.05em{\sc i\kern-.025em b}\kern-.08em
    T\kern-.1667em\lower.7ex\hbox{E}\kern-.125emX}}
\begin{document}

\title{SenseWeather: Multimodal Sensing for Robust Robotic Perception under Extreme Conditions
}

\author{
Fucai Ke$^{1}$,
Sukai Huang$^{1}$,
Junyi Hao$^{1}$,
Kal Backman$^{1}$,
Naim Rastgoo$^{1}$,
Gholamreza Haffari$^{1}$,
Hamid Rezatofighi$^{1}$\\
{\small $^{1}$Faculty of Information Technology, Monash University, Clayton, VIC 3800, Australia}
}

\maketitle

\begin{abstract}
Robotic perception must remain reliable when extreme environmental conditions severely degrade the primary RGB observation. Although robotic platforms commonly carry complementary sensors such as depth and infrared, much of the modern perception stack remains RGB-centric, creating a mismatch between multimodal onboard sensing and pretrained downstream models. We study whether complementary sensing can be used upstream to recover a task-compatible RGB observation that preserves existing perception capabilities without modifying or retraining the downstream stack. To this end, we introduce \emph{SenseWeather}, a controlled multimodal benchmark for robotic perception under severe rain, snow, dust, fog, and darkness, with registered RGB, depth, and infrared observations, clean references, and grounded high-level perception tasks. We further develop \emph{Fusion-to-RGB}, with two parameter-efficient sensor-fusion realizations based on token-space fusion for a diffusion transformer and residual latent conditioning for an inpainting UNet. Across frozen open-vocabulary grounding and vision-language models, RGB+D+IR fusion improves all paired downstream comparisons over the corresponding RGB-only restorers, including AP$_{50}$ gains across all ten detector comparisons. Importantly, these perception gains do not consistently follow conventional image-fidelity metrics, and increasing weather severity does not induce monotonic downstream degradation. These results show that complementary sensing can preserve reusable robotic perception under extreme conditions, while highlighting task-level perception robustness as an objective distinct from visual restoration quality.
\end{abstract}


\section{Introduction}\label{sec_intro}

\begin{figure*}[t]
    \centering
    \includegraphics[width=0.95\textwidth]{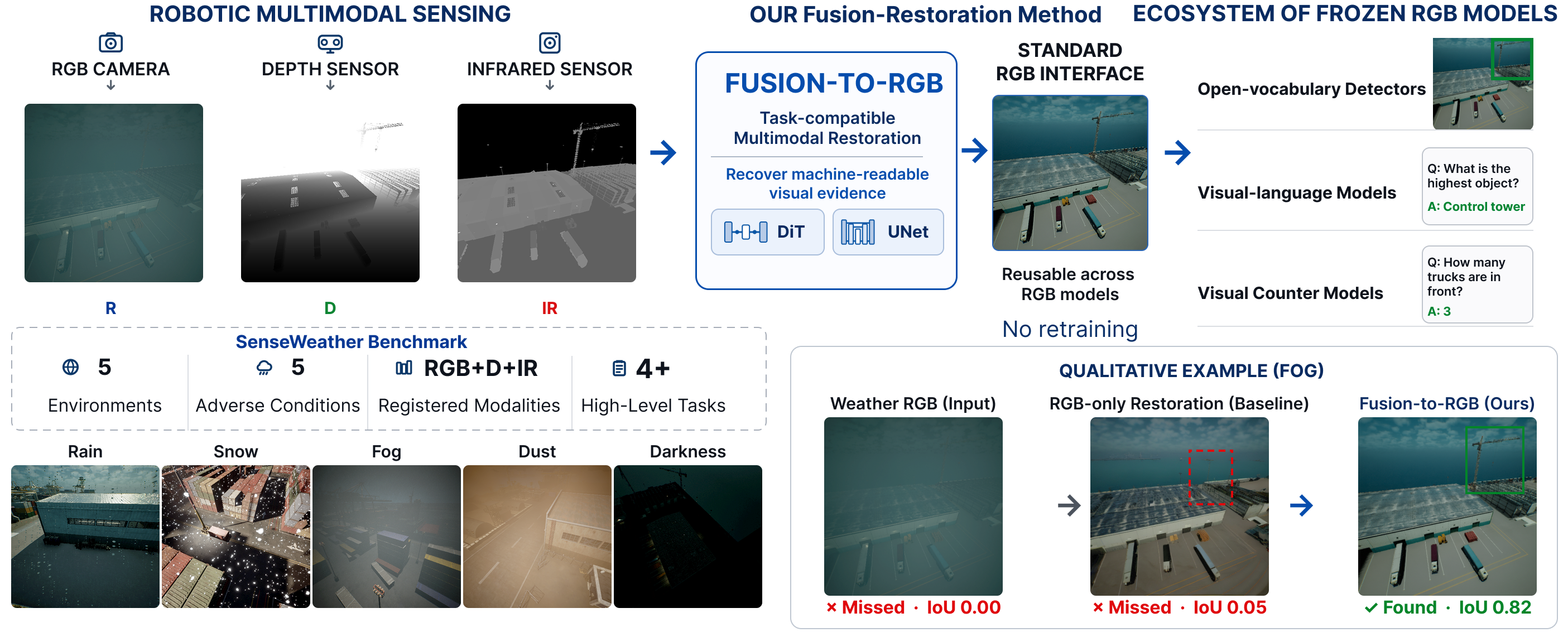}\vspace{-1em}
    \caption{SenseWeather studies Fusion-to-RGB as a task-compatible multimodal interface: complementary sensing recovers machine-readable RGB evidence for unchanged foundation models, even when RGB-only restoration remains ambiguous.}
    \label{fig:teaser}\vspace{-1.5em}
\end{figure*}

Modern robotic systems increasingly depend on RGB-centric perception as the front end of their perception--action pipeline. Advances in object detection, vision-language models, and embodied reasoning have enabled powerful pretrained models to support scene understanding, target localization, decision making, and interaction across a wide range of robotic platforms~\cite{pmlr-v205-ichter23a, An2023RGBManipMI, ke2026view2space, Qureshi2024SplatSimZS, ke2024hydra, Han2025MultimodalFAA, Li2022EmergentVSA, Sapkota2025VisionLanguageActionMA}. Reliable operation, however, requires these perception capabilities to remain functional when robots encounter extreme environmental conditions. Severe rain, fog, dust, snow, and poor illumination can corrupt the RGB stream, obscure scene structure, and remove task-relevant visual evidence~\cite{lin2024nightrain, lin2025nighthaze, du2026weatherreasonseg}. Such degradation is therefore not merely an image-quality problem: it can propagate through the robotic perception stack, causing failures in localization, scene understanding, and subsequent decision making.

Robotic platforms, however, typically operate with sensing redundancy rather than RGB alone. Depth cameras, infrared sensors, and other onboard modalities provide complementary geometric, spectral, and structural cues that can remain reliable when visual appearance is degraded~\cite{Bavle2021FromSTA, Palladin2025SAMFusionSMA, lin2026geocomplete, Guo2024DAEFuseAAA}. Depth can preserve scene geometry and object boundaries under poor visibility, while infrared can retain discriminative cues under challenging illumination. In principle, such redundancy should allow a robot to maintain perception when its RGB stream becomes unreliable. In practice, however, much of the modern perception stack is built around pretrained RGB models that cannot directly consume these auxiliary observations. This creates a deployment-level interface mismatch between the multimodal sensors available on the robot and the RGB-centric models used for perception, reasoning, and action~\cite{Bai2022TransFusionRLA, yang2026world, mazhar2021gem}.

Existing approaches offer two incomplete solutions to this deployment problem. RGB restoration preserves the standard visual interface, but typically relies on the corrupted RGB observation alone~\cite{Jiang2024ASOA}. When adverse conditions remove task-relevant evidence, such methods must reconstruct it without access to the complementary measurements already available on the robot~\cite{Zhang2024SSPIRSAA}. Task-specific multimodal perception takes the opposite route: it directly fuses RGB with depth, infrared, or other sensors~\cite{lin2026geocomplete, Guo2024DAEFuseAAA}, but usually requires dedicated architectures, supervision, or retraining for each downstream capability~\cite{Chen_unirestore, Deng2019DeepCNA}. This limits reuse of the large ecosystem of pretrained RGB perception and reasoning models already available for robotic systems. We therefore ask whether multimodal sensing can instead be used upstream to recover a standard RGB observation that preserves complementary sensor evidence while remaining directly consumable by existing RGB models without downstream modification or retraining.

Realizing this goal presents several challenges. First, learning sensor-conditioned restoration requires registered degraded RGB, auxiliary sensor observations, and clear RGB references from the same robotic viewpoint and scene state, which are difficult to acquire systematically across controlled extreme conditions. Second, onboard sensors provide heterogeneous evidence, with depth capturing geometric structure and infrared providing complementary spectral cues, which must be effectively integrated into a common RGB observation. Third, preserving visual appearance alone is insufficient: the recovered observation must retain the task-relevant evidence required by the robot's downstream perception stack. Improvements in conventional restoration metrics therefore need not translate into better localization, recognition, counting, or visual reasoning~\cite{Chen_unirestore, parmar2024img2imgturbo}. Evaluating such a system therefore requires measuring both image fidelity and the ability of unchanged pretrained RGB models to recover reliable perception from the restored observation.

To study robotic perception under extreme environmental conditions, we construct \emph{SenseWeather}, a multimodal benchmark spanning five controlled yet realistic simulated robotic environments. It provides registered adverse RGB, clear RGB, depth, and infrared observations under severe rain, snow, dust, fog, and darkness, including unseen environments and held-out viewpoints. SenseWeather further evaluates high-level perception capabilities commonly required by robotic systems through open-vocabulary grounding and vision-language reasoning, covering localization, counting, presence recognition, and localized object identification. The benchmark therefore measures whether task-relevant scene understanding can be preserved when the robot's primary RGB observation is severely compromised.

Building on SenseWeather, we introduce \emph{Fusion-to-RGB}, which uses complementary onboard sensing to recover a standard RGB observation for existing robotic perception models. We develop two architecturally distinct realizations: token-space sensor fusion for a diffusion transformer and residual latent conditioning for an inpainting UNet. Both preserve the pretrained generative prior and require only parameter-efficient adaptation, while producing RGB observations that can be directly consumed by unchanged downstream models. This allows robotic systems to exploit multimodal sensing under extreme conditions without constructing a task-specific fusion model for each perception capability. Across open-vocabulary grounding and vision-language reasoning, Fusion-to-RGB improves all paired downstream comparisons over the corresponding RGB-only restorers, including cases where conventional image-fidelity metrics do not improve.

The contributions of this work are summarized as follows:
\begin{itemize}

\item We formulate extreme-condition robotic perception as a task-compatible multimodal RGB recovery problem, where complementary onboard sensing is used to produce a reusable RGB observation for existing perception models without modifying the downstream stack.

\item We introduce \emph{SenseWeather}, a controlled multimodal benchmark for extreme-condition robotic perception, providing registered RGB, depth, and infrared observations across diverse operational environments together with grounded high-level perception tasks.

\item We develop two parameter-efficient realizations of Fusion-to-RGB, using token-space fusion and residual latent conditioning to translate heterogeneous sensor evidence into RGB observations compatible with frozen perception models.

\item We show that RGB+D+IR fusion improves all paired grounding and
vision-language comparisons over the corresponding RGB-only restoration, while conventional image fidelity does not reliably predict downstream robotic perception performance.

\end{itemize}




\section{Related Work}\label{sec_related}

\paragraph{Adverse-Weather Image Restoration.}
Image restoration has been extensively studied for adverse conditions such as rain~\cite{chen2024dual,lin2024nightrain} and haze~\cite{lin2025nighthaze, lin2025seeing}, with recent all-in-one methods targeting multiple degradations within a unified framework~\cite{Jiang2024ASOA}. Generative priors have further improved realistic restoration through diffusion-based reconstruction~\cite{Zhang2024SSPIRSAA, lin2026rgb, lin2026polarscale}, task-oriented objectives~\cite{Chen_unirestore}, and pretrained image generation or completion models~\cite{parmar2024img2imgturbo,tang2024realfill,wu2025qwenimage}. These methods, however, remain predominantly RGB-centric and must infer information lost under severe degradation from the corrupted image itself. We instead use complementary sensing to recover the RGB observation.

\paragraph{Multimodal Sensor Fusion.}
Robotic platforms commonly combine RGB cameras with depth, infrared, or LiDAR sensors that provide complementary geometric and spectral cues~\cite{Han2025MultimodalFAA,ke2024hydra,Li2022EmergentVSA,ke2024hitskt,Bavle2021FromSTA}. Such information has been exploited through task-specific multimodal fusion for 3D detection and robust perception~\cite{Bai2022TransFusionRLA,ke2025explain,mazhar2021gem,Palladin2025SAMFusionSMA}, as well as through multimodal image fusion~\cite{Deng2019DeepCNA,Ke_2025_dwim,Guo2024DAEFuseAAA}. Most existing approaches produce task-specific predictions or fused representations. In contrast, we fuse sensing upstream to restore a standard RGB image, allowing heterogeneous pretrained RGB models to operate unchanged.

\paragraph{Restoration for Machine Perception.}
Restoration is traditionally evaluated using full-reference fidelity metrics such as PSNR, SSIM~\cite{wang2004ssim}, and LPIPS~\cite{zhang2018lpips}. These metrics measure similarity to a clean reference but do not necessarily reflect the usefulness of restored images to machine perception. Recent work has begun incorporating task-oriented restoration objectives~\cite{Chen_unirestore}, while adverse-weather benchmarks expose persistent degradation in modern vision and vision-language models~\cite{du2026weatherreasonseg}. Our evaluation instead keeps heterogeneous downstream models frozen and measures whether one restored RGB image transfers across detection and vision--language tasks. We therefore treat visual fidelity and machine readability as complementary, rather than interchangeable, objectives.
\section{SenseWeather Benchmark}\label{sec_benchmark}

\subsection{Benchmark Formulation}

SenseWeather studies whether robotic perception can remain reliable when extreme environmental conditions severely compromise the primary RGB observation. We consider a robot equipped with registered RGB, depth, and infrared sensors, while its downstream perception stack consists of pretrained RGB models. Given a degraded RGB observation $I^{adv}_{rgb}$ and complementary depth and infrared measurements $(D, I_{ir})$, a restoration model produces a conventional RGB observation
\begin{equation}
    \hat{I}_{rgb} = R_{\theta}(I^{adv}_{rgb}, D, I_{ir}),
    \qquad
    \hat{y}_{k} = F_k(\hat{I}_{rgb}),
\end{equation}
where $F_k$ denotes a frozen RGB model for downstream perception task $k$. A paired clear observation $I^{clr}_{rgb}$ is used only for restoration supervision and full-reference evaluation, and is never available at inference time.

This formulation separates multimodal sensing from downstream perception. Rather than constructing a dedicated multimodal model for each capability, the sensor-conditioned restorer produces one standard RGB observation that can be reused by heterogeneous frozen perception models. RGB-only restoration is recovered as the special case in which $D$ and $I_{ir}$ are omitted. SenseWeather therefore evaluates whether complementary onboard sensing preserves task-relevant robotic perception under extreme conditions, rather than whether a method is optimized for a particular downstream task.

\subsection{Extreme-Condition Environments and Multimodal Capture}

Collecting registered multimodal observations with paired clear references and controlled adverse conditions is difficult to achieve systematically in the real world. We therefore construct SenseWeather in AirSim~\cite{shah2018airsim} across five operationally distinct robotic environments: an island base, three offshore docks, and an oil refinery. These environments cover dense container terminals, industrial facilities, coastal terrain, vehicles, material-handling equipment, and maritime assets, supporting both UAV and ground-robot perception scenarios. Each environment is equipped with a curated semantic catalog that groups task-level objects into coarse and fine-grained categories. Across all five environments, the catalogs contain 22,495 object instances spanning 175 coarse and 247 fine-grained categories (Table~\ref{tab:scene_inventory}).

\begin{table*}[t]
    \centering
    \caption{Object inventory of the five SenseWeather environments before exclusion and visibility filtering. Each instance denotes one annotated object; coarse and fine-grained categories correspond to hierarchical labels (e.g., Architecture and Architecture-Hangar). The final row reports totals with categories deduplicated across environments.}
    \label{tab:scene_inventory}
    \resizebox{\textwidth}{!}{%
    \begin{tabular}{l l r r r}
        \hline
        \textbf{Environment} & \textbf{Representative operational content} & \textbf{Object instances} & \textbf{Coarse categories} & \textbf{Fine-grained categories} \\
        \hline
        Island Base & Coastal terrain, fortifications, vehicles, rotorcraft, and port equipment & 725 & 51 & 64 \\
        Offshore Dock 1 & Dense containers, cranes, trucks, reachstackers, forklifts, and utilities & 18,488 & 61 & 90 \\
        Offshore Dock 2 & Ship-handling terminal, STS cranes, carriers, barriers, and service assets & 1,059 & 52 & 65 \\
        Offshore Dock 3 & Cargo storage, tanks, fences, trailers, trucks, and material handling & 1,408 & 29 & 41 \\
        Oil Refinery & Process pipes, storage facilities, industrial frames, cranes, and vessels & 815 & 16 & 21 \\
        \hline
        \textbf{All environments} & Five environment-specific catalogs & \textbf{22,495} & \textbf{175} & \textbf{247} \\
        \hline
    \end{tabular}%
    }\vspace{-1.5em}
\end{table*}

Observations are captured from UAV- and robot-centric viewpoints around annotated scene objects. Each accepted viewpoint provides registered $512\!\times\!512$ clear RGB, degraded RGB, depth, infrared, and instance annotations. We consider severe rain, snow, dust, and fog together with a separate darkness condition. Visibility and framing checks remove views with poorly observed or clipped targets, while paired capture preserves the same camera pose and scene state across sensing conditions. This produces spatially aligned multimodal observations suitable for both restoration and downstream robotic perception evaluation. Detailed sampling and capture procedures are provided in the supplementary material.

We use approximately 10k paired observations for restoration training and construct a manually curated \emph{Primary Extreme-Condition Test Set} of 625 observations for downstream evaluation. It contains 500 observations under severe rain, snow, dust, and fog at intensities $0.8$, $0.9$, and $1.0$, together with 125 registered darkness observations from non-reused viewpoints. The test set covers four environments unseen during training and held-out viewpoints from Offshore Dock~1, providing both cross-environment and cross-view evaluation. In total, it supports 625 localization queries with 825 target boxes and 939 vision-language questions.

We additionally construct a \emph{Controlled Intensity Study Set} with 400 manually reviewed observations from 10 fixed viewpoints. For each viewpoint, scene state and sensing configuration are held fixed while rain, snow, dust, and fog intensity varies from $0.1$ to $1.0$. This controlled split isolates the effect of weather severity on robotic perception from changes in scene composition; darkness is evaluated separately as a fixed extreme illumination condition.

\subsection{Grounded Robotic Perception Tasks and Evaluation}

SenseWeather evaluates whether observations under extreme conditions retain the spatial and semantic evidence required for high-level robotic scene understanding. We consider open-vocabulary localization together with three vision-language capabilities: category counting, object presence reasoning, and localized object identification. All tasks are instantiated directly from the recorded simulator state and subsequently audited, ensuring that target labels, questions, answers, and spatial groundings remain consistent with the observed scene. Localization prompts cover diverse object categories, while the vision--language tasks probe complementary aspects of scene understanding beyond object detection alone.

Our primary evaluation measures task-level perception using frozen pretrained models. Open-vocabulary localization is evaluated with AP$_{50}$ and COCO-style mAP$_{50:95}$; category counting uses exact-match accuracy, presence reasoning uses balanced accuracy, and localized object identification uses four-way choice accuracy. For every comparison, the target annotation, prompt, scoring rule, and downstream model are fixed, while only the RGB observation supplied to the model changes among degraded, restored, and clear inputs. This controlled image-substitution protocol directly measures whether restoration recovers task-relevant evidence without adapting the downstream perception model.

Full-reference metrics, including PSNR, SSIM~\cite{wang2004ssim}, LPIPS~\cite{zhang2018lpips}, and RGB MAE, are reported separately on paired weather observations as diagnostic measures of visual fidelity. They are not used as proxies for robotic perception performance, allowing us to explicitly examine when image similarity and downstream perception diverge.



\section{Sensor-Conditioned Fusion-to-RGB Restoration}
\label{sec:fusion_method}

\subsection{Task-Compatible Fusion-to-RGB}
Fusion-to-RGB separates multimodal sensing from downstream task inference. Given a degraded RGB observation $I_{\mathrm{rgb}}^{\mathrm{adv}}$ and its registered depth and infrared measurements $(D,I_{\mathrm{ir}})$, the restoration model $R_\theta$ produces a conventional RGB observation $\widehat I_{\mathrm{rgb}}$, which is directly consumed by a frozen downstream RGB model $F_k$:
\begin{equation}
    \widehat I_{\mathrm{rgb}}
    =
    R_\theta\!\left(
    I_{\mathrm{rgb}}^{\mathrm{adv}},D,I_{\mathrm{ir}}
    \right),
    \qquad
    \widehat y_k
    =
    F_k\!\left(\widehat I_{\mathrm{rgb}}\right).
    \label{eq:rgb_interface}
\end{equation}
Here, $k$ indexes the downstream perception task. The paired clear image $I_{\mathrm{rgb}}^{\mathrm{clr}}$ provides supervision during training but is never available at inference. This formulation allows a single restored RGB observation to be reused across heterogeneous frozen perception models, separating sensor fusion from task-specific inference.

\begin{figure*}[t]
    \centering
    \includegraphics[width=\textwidth]{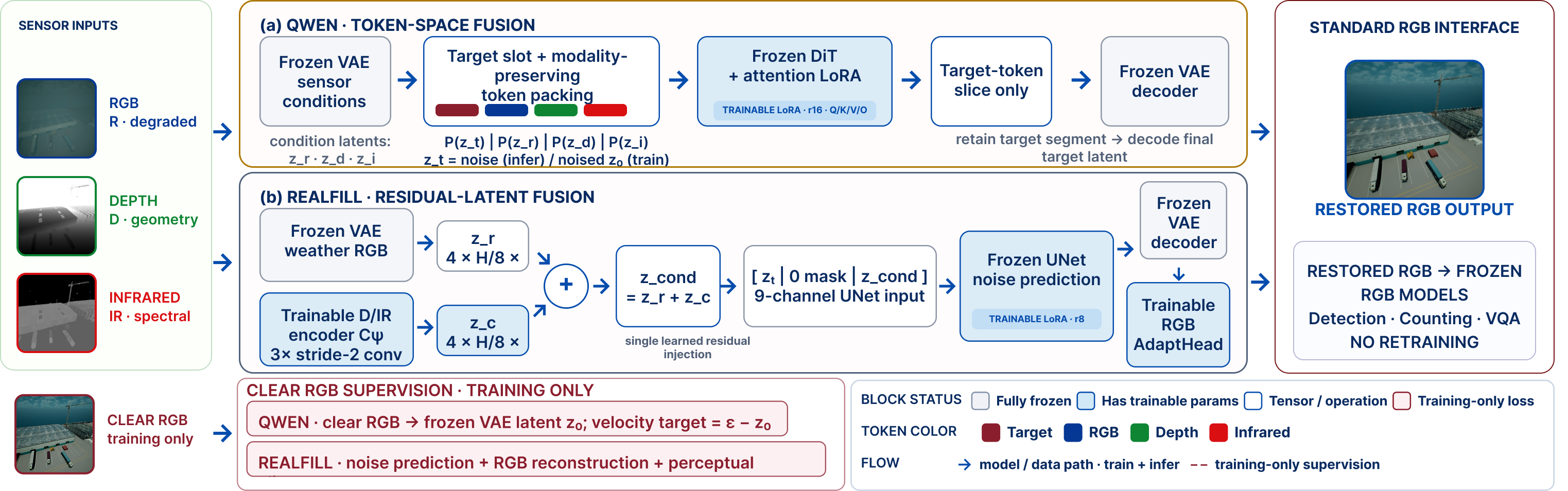}
    \caption{Fusion-to-RGB preserves one downstream interface while adapting fusion to the backbone. Qwen keeps the modalities as separate image-token segments; RealFill injects a learned depth--infrared residual into its RGB conditioning latent. The clear target appears only during training, and every downstream model remains frozen.}
    \vspace{-1em}
    \label{fig:fusion_architecture}
\end{figure*}

We build Fusion-to-RGB on pretrained diffusion models and restrict adaptation to lightweight trainable components. This retains the pretrained generative prior while introducing geometric and spectral evidence from depth and infrared that can remain informative when RGB is severely degraded. We realize this principle with two architecturally distinct designs (Fig.~\ref{fig:fusion_architecture}): token-space fusion for Qwen-Image-Edit~\cite{wu2025qwenimage} and residual latent conditioning for the UNet-based RealFill backbone~\cite{tang2024realfill}. Both produce the same conventional RGB output for unchanged downstream models, while integrating sensor evidence at a location suited to the underlying architecture.


\subsection{Token-Space Fusion for Diffusion Transformers}

Let $E$ denote the frozen Qwen VAE and $P$ its latent packing operation. We independently encode the clear target $z_0=E(I_{\mathrm{rgb}}^{\mathrm{clr}})$ and the three inference-time conditions $z_r=E(I_{\mathrm{rgb}}^{\mathrm{adv}})$, $z_d=E(D)$, and $z_i=E(I_{\mathrm{ir}})$. For a flow-matching timestep $t$, only the target latent is corrupted:
\begin{equation}
    z_t=(1-\sigma_t)z_0+\sigma_t\epsilon, \qquad
    H_t=[P(z_t)\mid P(z_r)\mid P(z_d)\mid P(z_i)].
    \label{eq:qwen_fusion_tokens}
\end{equation}

The transformer jointly processes all four latent segments while retaining their individual spatial metadata, preserving modality boundaries while allowing self-attention to exchange information across the registered observations. Let $T_{\Phi,\Delta}$ denote the transformer with frozen base parameters $\Phi$ and trainable attention LoRA parameters $\Delta$. We retain the output corresponding to the target segment, unpack it to the target latent layout, and compute the flow-matching objective:
\begin{equation}
    \widehat v_t =
    U\!\left(
    T_{\Phi,\Delta}(H_t,t,c)_{1:|P(z_t)|}
    \right),
    \qquad
    \mathcal L_{\mathrm{Q}}
    =
    \left\|
    \widehat v_t-(\epsilon-z_0)
    \right\|_2^2,
    \label{eq:qwen_fusion_loss}
\end{equation}
where $U$ denotes the latent unpacking operation and $c$ is the restoration instruction. The VAE, prompt encoders, and base transformer remain frozen, with LoRA applied to the query, key, value, and output attention projections. At inference, the adverse RGB, depth, and infrared condition latents remain fixed throughout denoising and are concatenated with the evolving target latent at every step. The transformer jointly processes all four segments, while only the target prediction is retained to update the target latent. After the final denoising step, the target latent alone is decoded to produce the restored RGB observation.

\subsection{Residual-Latent Fusion for UNet Diffusion}

RealFill operates on spatial conditioning latents rather than separate image-token segments. We retain its frozen VAE $E$ and encode the degraded RGB observation as $z_r=E(I_{\mathrm{rgb}}^{\mathrm{adv}})$. A lightweight trainable sensor encoder $C_\psi$ maps the registered depth and infrared observations to a four-channel latent residual $z_c$ with the same spatial resolution as $z_r$. Its final projection is zero-initialized, so the model begins from the pretrained RGB-conditioning pathway:
\begin{equation}
    z_c=C_\psi(D,I_{\mathrm{ir}}), \qquad
    z_{\mathrm{cond}}=z_r+z_c.
    \label{eq:realfill_latent_fusion}
\end{equation}
The inpainting UNet receives the channel-wise concatenation of the noisy target latent, an all-zero mask, and $z_{\mathrm{cond}}$. We freeze the VAE, text encoder, and base UNet parameters, while training attention LoRA in the UNet together with $C_\psi$ and a lightweight residual RGB adaptation head $A_\omega$. The diffusion branch follows the standard noise-prediction objective. At each sampled timestep, we recover an estimate of the clean latent from the predicted noise and decode it with the frozen VAE to obtain $\widetilde I$. The adaptation head then applies a residual correction in pixel space,
\begin{equation}
    \widehat I_{\mathrm{rgb}} = A_\omega(\widetilde I).
\end{equation}
Training combines diffusion supervision with paired RGB reconstruction and perceptual alignment:
\begin{equation}
    \mathcal L_{\mathrm{R}}
    =
    \mathcal L_{\mathrm{diff}}
    +\lambda_{\mathrm{pix}}
    \|\widehat I_{\mathrm{rgb}}-I_{\mathrm{rgb}}^{\mathrm{clr}}\|_1
    +\lambda_{\mathrm{perc}}
    \mathcal L_{\mathrm{perc}}
    (\widehat I_{\mathrm{rgb}},I_{\mathrm{rgb}}^{\mathrm{clr}}).
    \label{eq:realfill_fusion_loss}
\end{equation}
Unlike multi-resolution control architectures, the auxiliary sensors contribute a single learned residual to the RGB conditioning latent. This provides lightweight sensor conditioning while preserving the pretrained UNet pathway.

\section{Experiments and Analysis}
\label{sec_experiments}

Our experiments address three questions. First, can complementary sensing preserve high-level robotic perception under extreme conditions while keeping the downstream perception stack unchanged? Second, how do depth and infrared contribute across restoration backbones, perception tasks, and environmental conditions? Third, do conventional image-fidelity metrics and commanded weather severity provide reliable proxies for downstream perception robustness? We answer these questions through controlled image substitution with frozen downstream models.

\subsection{Experimental Setup}

\paragraph{Restoration models.}
We compare four RGB-only restoration baselines with six sensor-conditioned variants of Fusion-to-RGB. The RGB-only baselines include MAE~\cite{he2022mae}, Pix2Pix-Turbo~\cite{parmar2024img2imgturbo}, Qwen-Image-Edit~\cite{wu2025qwenimage}, and RealFill~\cite{tang2024realfill}. For Qwen and RealFill, we further evaluate RGB+D, RGB+IR, and RGB+D+IR variants to isolate the contribution of complementary sensing. All learned models are trained on the same approximately 10k paired observations from Offshore Dock~1, with degraded RGB as input, optional registered depth and infrared as auxiliary observations, and clear RGB as supervision. The training data are disjoint from all evaluation sets, and model selection and inference settings are fixed before evaluation.

\paragraph{Frozen downstream perception.}
We evaluate restored observations with five open-vocabulary grounding models, Grounding DINO, Florence-2, Falcon-Perception-300M, OmDet-Turbo, and YOLO-World~\cite{liu2024groundingdino,xiao2024florence2,bevli2026falcon,zhao2024omdetturbo,cheng2024yoloworld}, together with Qwen3-VL-2B and Qwen3-VL-4B~\cite{bai2025qwen3vl}. All downstream models remain frozen. For each paired comparison, the target annotation, prompt, scoring protocol, model weights, and inference settings are held fixed; only the RGB observation supplied to the model is changed. This controlled image-substitution protocol directly tests whether complementary sensing recovers task-relevant evidence that can be reused by an unchanged robotic perception stack.

\paragraph{Evaluation splits.}
Unless otherwise stated, downstream perception is evaluated on the primary extreme-condition test set described in Sec.~\ref{sec_benchmark}. Full-reference metrics are reported separately on observations with paired clear RGB and are used only to analyze visual fidelity rather than as a proxy for perception performance. The Controlled Intensity Study Set is reserved for the severity analysis in Sec.~\ref{sec:intensity_analysis}.

\subsection{Preserving Robotic Perception under Extreme Conditions}

\begin{figure}[!ht]
    \centering
    \includegraphics[width=\columnwidth]{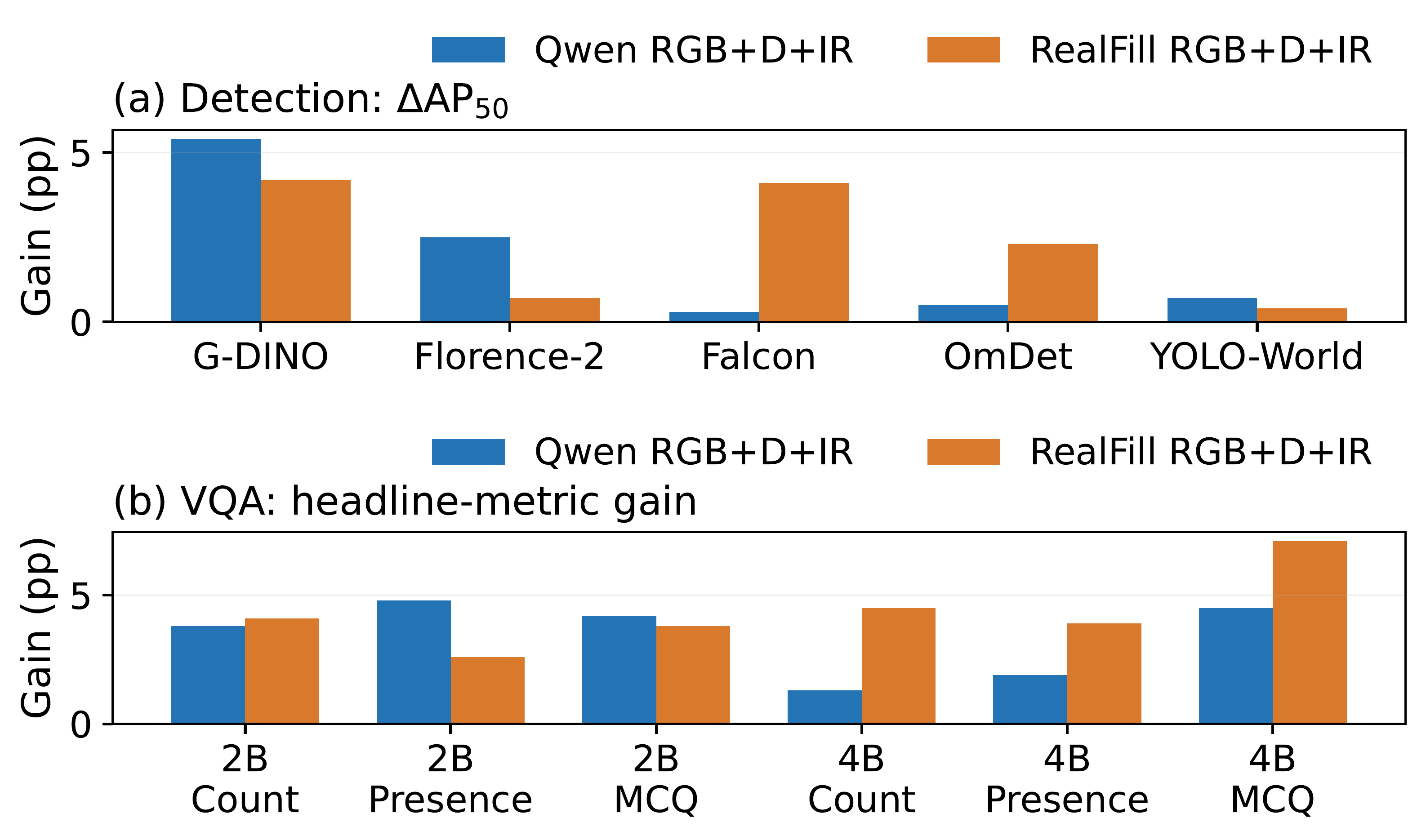}
    \caption{RGB+D+IR fusion improves AP$_{50}$ in all ten paired detector comparisons and improves all twelve vision--language comparisons. The plot shows percentage-point gains over each RGB-only parent for AP$_{50}$ (top) and the VLM headline metrics (bottom).}
    \vspace{-1em}
    \label{fig:paired_fusion_gain}
\end{figure}

We first ask whether complementary sensing improves perception across heterogeneous frozen models, rather than benefiting a particular downstream task or generative backbone. Across both Fusion-to-RGB realizations, RGB+D+IR improves every paired comparison with its RGB-only counterpart: all ten open-vocabulary localization evaluations and all twelve vision-language evaluations (Fig.~\ref{fig:paired_fusion_gain}). This consistency indicates that the recovered sensor evidence transfers across downstream perception models without task-specific adaptation.

\subsubsection{Open-Vocabulary Localization}

RGB+D+IR improves AP$_{50}$ for all five detectors with both Qwen and RealFill (Tables~\ref{tab:localization} and~\ref{tab:localization_fusion}). The gains range from $+0.5$ to $+5.4$ percentage points for Qwen and from $+0.7$ to $+4.2$ points for RealFill. Despite substantial differences among detector architectures, both restoration backbones show the same positive trend, suggesting that complementary sensing recovers localization evidence that is reusable across heterogeneous perception models.

\begin{table}[t]
    \centering
    \caption{Performance of RGB-only restoration baselines on open-vocabulary localization under extreme conditions. Each entry is AP$_{50}$ (\%). Bold marks the best learned restoration for each detector. Florence-2 and Falcon assign tied confidence to retained boxes, so their AP should only be compared across image domains within the same detector.}
    \label{tab:localization}
    \resizebox{\columnwidth}{!}{%
    \begin{tabular}{lccccc}
        \toprule
        \textbf{Image domain} & \textbf{Grounding DINO} & \textbf{Florence-2} & \textbf{Falcon} & \textbf{OmDet-Turbo} & \textbf{YOLO-World} \\
        \midrule
        \rowcolor{SWReference} Degraded RGB & 30.9 & 15.0 & 3.4 & 9.8 & 5.7 \\
        MAE~\cite{he2022mae} & 11.5 & 4.9 & \textbf{4.4} & 2.8 & 1.0 \\
        Pix2Pix-Turbo~\cite{parmar2024img2imgturbo} & 28.4 & 13.2 & 4.0 & 8.1 & 4.4 \\
        RealFill~\cite{tang2024realfill} & 26.3 & 9.3 & 2.0 & 8.2 & 3.6 \\
        Qwen-image~\cite{wu2025qwenimage} & \textbf{35.7} & \textbf{18.3} & 4.0 & \textbf{12.6} & \textbf{6.6} \\
        \rowcolor{SWReference} Ideal RGB & 42.1 & 24.2 & 4.0 & 14.1 & 7.5\\
        \bottomrule
    \end{tabular}}
\end{table}

\begin{table}[t]
    \centering
    \caption{Effect of complementary sensing in our Fusion-to-RGB framework on open-vocabulary localization. RGB denotes the RGB-only baseline for each backbone, while RGB+D, RGB+IR, and RGB+D+IR are sensor-conditioned Fusion-to-RGB variants. Each entry reports AP$_{50}$ (\%) on the Primary Extreme-Condition Test Set. $\Delta$ denotes RGB+D+IR minus the RGB baseline in percentage points.}
    \label{tab:localization_fusion}
    \resizebox{\columnwidth}{!}{%
    \begin{tabular}{llccccc}
        \toprule
        \textbf{Backbone} & \textbf{Input} 
        & \textbf{Grounding DINO} 
        & \textbf{Florence-2} 
        & \textbf{Falcon} 
        & \textbf{OmDet-Turbo} 
        & \textbf{YOLO-World} \\
        \midrule

        \multirow{5}{*}{RealFill~\cite{tang2024realfill}}
        & RGB        & 26.3 & 9.3 & 2.0 & 8.2 & 3.6 \\
        & RGB+D      & 30.4 & 9.8 & 4.0 & 9.4 & 4.4  \\
        & RGB+IR      & 29.0 & 8.9 & 4.9 & 8.6 & 4.1 \\
        & \textbf{RGB+D+IR} 
                     & \textbf{30.5} & \textbf{10.0} & \textbf{6.1} & \textbf{10.5} & \textbf{4.8} \\
        & $\Delta$   & \textcolor{red!75!black}{+4.2}
                     & \textcolor{red!75!black}{+0.7}
                     & \textcolor{red!75!black}{+4.1}
                     & \textcolor{red!75!black}{+2.3}
                     & \textcolor{red!75!black}{+1.2} \\

        \midrule

        \multirow{5}{*}{Qwen-image~\cite{wu2025qwenimage}}
        & RGB        & 35.7 & 18.3 & 4.0 & 12.6 & 6.6 \\
        & RGB+D     & 35.9 & 18.9 & 4.6 & 12.5 & 6.6 \\
        & RGB+IR     & 38.1 & 19.4 & 5.3 & 12.8 & 6.8 \\
        & \textbf{RGB+D+IR}
                     & \textbf{41.1} & \textbf{20.8} & \textbf{5.6} & \textbf{13.1} & \textbf{7.3} \\
        & $\Delta$   & \textcolor{red!75!black}{+5.4}
                     & \textcolor{red!75!black}{+2.5}
                     & \textcolor{red!75!black}{+1.6}
                     & \textcolor{red!75!black}{+0.5}
                     & \textcolor{red!75!black}{+0.7} \\

        \bottomrule
    \end{tabular}}
\end{table}


\begin{figure*}[t]
    \centering
    \includegraphics[width=0.9\textwidth]{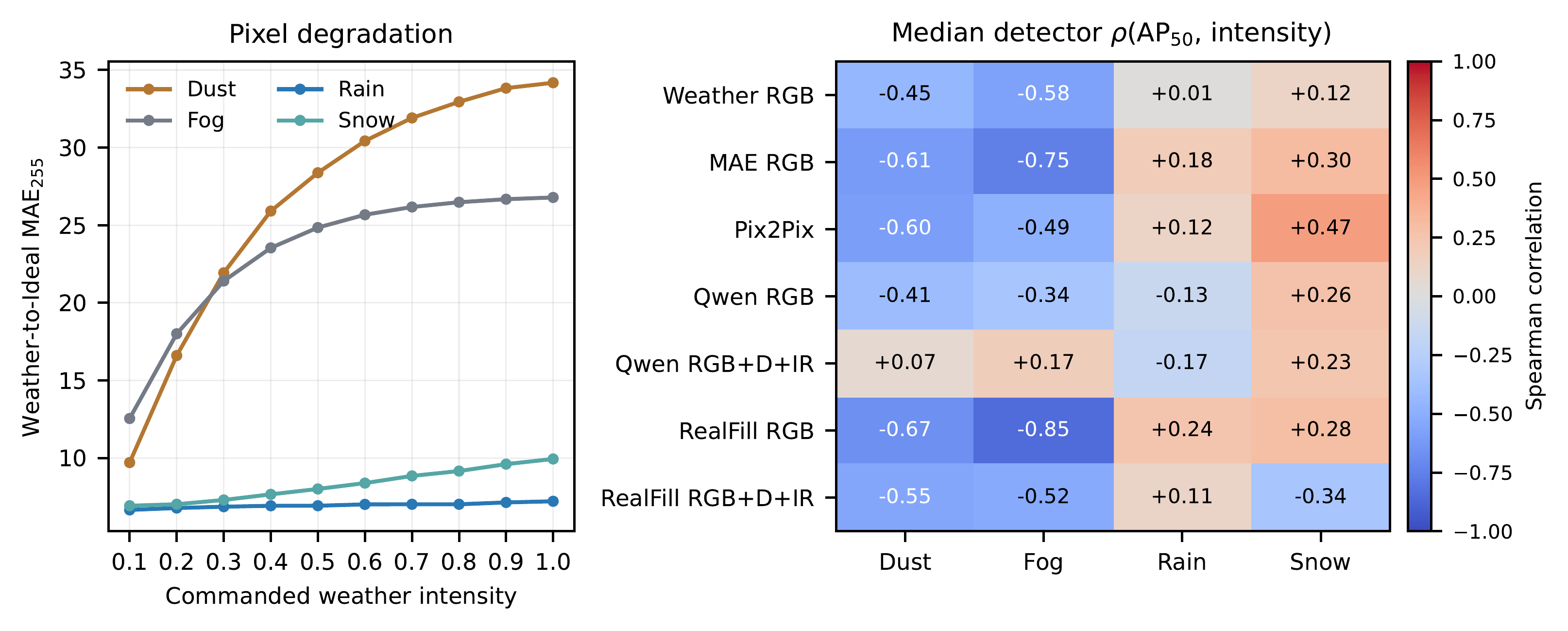}
    \vspace{-1em}
    \caption{Weather intensity produces monotonic pixel degradation but not a monotonic detection response on the Controlled Intensity Study Set. Left: MAE increases within every weather, with much larger absolute effects for dust and fog. Right: median Spearman correlation between intensity and AP$_{50}$ across five detectors for each image domain; negative is the expected degradation direction. Ideal RGB is omitted because it is an input reference rather than a restoration method. Darkness is excluded because it was added only to the expanded primary evaluation and has no controlled intensity sweep.}
    \vspace{-1em}
    \label{fig:intensity_analysis}
\end{figure*}

\subsubsection{Vision-Language Tasks}
The benefit extends beyond localization to semantic scene understanding. RGB+D+IR improves all twelve paired comparisons across Qwen3-VL model scales and the three evaluated capabilities: counting, presence reasoning, and localized object identification (Tables~\ref{tab:vqa} and~\ref{tab:vqa_fusion}). Gains range from $+1.3$ to $+7.1$ percentage points, showing that complementary sensing recovers not only spatial cues for object grounding but also visual evidence used for higher-level reasoning.

Two observations further illustrate the dependence on the downstream evaluator. Qwen RGB+D+IR reaches $76.4\%$ accuracy on 4B localized recognition, slightly exceeding the $75.7\%$ obtained with Ideal RGB; the clear image is therefore an input reference rather than a strict performance upper bound for an imperfect model. Moreover, increasing Qwen3-VL from 2B to 4B improves localized recognition but reduces exact counting across image domains. These results reinforce the need to evaluate perception robustness across multiple capabilities and model families rather than through a single downstream evaluator.


\begin{table}[t]
    \centering
    \caption{RGB-only restoration exhibits task- and scale-dependent utility for Qwen3-VL. All entries are percentages, and bold marks the best learned restoration in each column.}
    \label{tab:vqa}
    \resizebox{\columnwidth}{!}{%
    \begin{tabular}{lccc ccc}
        \toprule
        & \multicolumn{3}{c}{\textbf{Qwen3-VL-2B}} & \multicolumn{3}{c}{\textbf{Qwen3-VL-4B}} \\
        \cmidrule(lr){2-4}\cmidrule(lr){5-7}
        \textbf{Image domain} & \textbf{Count} & \textbf{Presence} & \textbf{MCQ} & \textbf{Count} & \textbf{Presence} & \textbf{MCQ} \\
        \midrule
        \rowcolor{SWReference} Degraded RGB & 45.7 & 73.8 & 56.5 & 30.4 & 74.2 & 65.8 \\
        MAE RGB & 40.3 & 70.0 & 45.7 & 24.9 & 68.4 & 50.5 \\
        Pix2Pix-Turbo & \textbf{51.8} & \textbf{70.6} & 59.7 & 33.5 & 71.0 & 62.9 \\
        RealFill RGB & 45.7 & 69.6 & 60.4 & 33.5 & 72.2 & 62.9 \\
        Qwen RGB & 50.5 & 67.1 & \textbf{66.1} & \textbf{41.5} & \textbf{72.2} & \textbf{71.9} \\
        \rowcolor{SWReference} Ideal RGB & 55.6 & 73.5 & 74.1 & 48.2 & 76.4 & 75.7 \\
        \bottomrule
    \end{tabular}}\vspace{-1em}
\end{table}

\begin{table}[t]
    \centering
\caption{Effect of complementary sensing in our Fusion-to-RGB framework on frozen Qwen3-VL performance. RGB is the RGB-only baseline for each backbone; RGB+D, RGB+IR, and RGB+D+IR are our sensor-conditioned variants. Results use the Primary Extreme-Condition Test Set. The $\Delta$ row reports RGB+D+IR minus RGB in percentage points.}
    \label{tab:vqa_fusion}
    \resizebox{\columnwidth}{!}{%
    \begin{tabular}{lcccc ccc}
        \toprule
        & & \multicolumn{3}{c}{\textbf{Qwen3-VL-2B}} & \multicolumn{3}{c}{\textbf{Qwen3-VL-4B}} \\
        \cmidrule(lr){3-5}\cmidrule(lr){6-8}
        \textbf{Backbone} & \textbf{Input} & \textbf{Count} & \textbf{Presence} & \textbf{MCQ} & \textbf{Count} & \textbf{Presence} & \textbf{MCQ} \\

        \midrule
        \multirow{5}{*}{RealFill~\cite{tang2024realfill}} & RGB & 45.7 & 69.6 & 60.4 & 33.5 & 72.2 & 62.9 \\
        & RGB+D & 50.5 & 69.6 & 63.6 & 36.1 & 73.8 & 68.4 \\
        & RGB+IR & 47.6 & 68.3 & 60.4 & 37.4 & 72.2 & 67.4 \\
        & RGB+D+IR & 49.8 & 72.2 & 64.2 & 38.0 & 76.1 & 70.0 \\
        &  $\Delta$  & \textcolor{red!75!black}{+4.1} & \textcolor{red!75!black}{+2.6} & \textcolor{red!75!black}{+3.8} & \textcolor{red!75!black}{+4.5} & \textcolor{red!75!black}{+3.9} & \textcolor{red!75!black}{+7.1} \\
        
        \midrule
        \multirow{5}{*}{Qwen-image~\cite{wu2025qwenimage}}
        & RGB & 50.5 & 67.1 & 66.1 & 41.5 & 72.2 & 71.9 \\
        & RGB+D & 54.0 & 69.6 & 66.8 & 38.7 & 77.3 & 72.2 \\
        & RGB+IR & 51.8 & 68.3 & 70.0 & 41.5 & 73.8 & 74.8 \\
        & RGB+D+IR & 54.3 & 71.9 & 70.3 & 42.8 & 74.1 & 76.4 \\
        & $\Delta$  & \textcolor{red!75!black}{+3.8} & \textcolor{red!75!black}{+4.8} & \textcolor{red!75!black}{+4.2} & \textcolor{red!75!black}{+1.3} & \textcolor{red!75!black}{+1.9} & \textcolor{red!75!black}{+4.5} \\
        \bottomrule
    \end{tabular}}\vspace{-1em}
\end{table}


\subsection{How Do Complementary Sensors Contribute?}

The benefit of auxiliary sensing is not attributable to a single dominant modality. In open-vocabulary localization, the relative value of depth and infrared depends strongly on the generative backbone. For Qwen, RGB+IR outperforms RGB+D across all five detectors, whereas RealFill benefits more from depth in four of the five comparisons (Table~\ref{tab:localization_fusion}). Despite this backbone dependence, combining both modalities is consistently strongest: RGB+D+IR achieves the best localization result for every backbone--detector pair.

A similar pattern appears in vision--language reasoning (Table~\ref{tab:vqa_fusion}). Neither depth nor infrared alone dominates across model scales and capabilities, while RGB+D+IR achieves the best result in ten of the twelve comparisons and improves over the RGB-only parent in all twelve. These results suggest that depth and infrared provide complementary evidence whose usefulness varies with the restoration architecture and downstream task. Joint conditioning therefore offers a more robust strategy than relying on either auxiliary modality alone.

\subsection{Perception Gains beyond Visual Fidelity}

Across their respective evaluation splits, conventional restoration metrics do not reliably predict downstream perception. On the primary extreme-condition test set, Qwen RGB+D+IR outperforms Qwen RGB across all five detectors and all six vision--language comparisons. On the separate paired-weather fidelity split, however, Qwen RGB+D+IR has lower PSNR ($13.40$ vs.~$14.27$) and SSIM ($0.537$ vs.~$0.578$), and higher MAE ($50.61$ vs.~$42.93$). RealFill exhibits a different regime: RGB+D+IR improves downstream perception while also improving fidelity, increasing PSNR from $14.02$ to $14.82$ and reducing MAE from $44.85$ to $41.81$. These contrasting behaviors show that gains in robotic perception do not consistently track improvements in pixel-level similarity to the clear reference. We therefore treat full-reference fidelity as a diagnostic measure rather than a proxy for task-level perception robustness.

\subsection{When Does Multimodal Sensing Help?}

\begin{table}[t]
    \centering
    \caption{Condition-dependent gain of RGB+D+IR over RGB-only restoration. Values are mean percentage-point improvements across Grounding DINO and VQA for the Qwen and RealFill families.}\vspace{-1em}
    \label{tab:condition_triple_fusion}
    \begin{tabular}{lccccc}
        \toprule
        \textbf{Condition} & Fog & Dust & Darkness & Rain & Snow\\
        \midrule
        \textbf{Mean gain (pp)}  & \textbf{+7.11} & \textbf{+6.43} & \textbf{+4.33} & +0.85 & $-0.10$ \\
        \bottomrule
    \end{tabular}
    \vspace{-1em}
\end{table}
The benefit of RGB+D+IR fusion varies substantially across environmental conditions (Table~\ref{tab:condition_triple_fusion}). For each condition, we average four RGB+D+IR-minus-RGB gains: Grounding DINO AP$_{50}$ and the mean VQA gain for each of the Qwen and RealFill backbones. Fusion yields the largest improvements under fog ($+7.11$ pp), dust ($+6.43$ pp), and darkness ($+4.33$ pp), while gains are modest under rain ($+0.85$ pp) and negligible under snow ($-0.10$ pp). We use this average only to summarize the condition-level trend.

The smaller gains under rain and snow are consistent with their weaker effective RGB degradation in our simulator: the controlled intensity sweep produces substantially smaller weather-to-clear MAE changes for these conditions than for fog and dust. We therefore attribute the limited gains primarily to the current degradation regime rather than to an inherent limitation of multimodal sensing. Darkness is excluded from this comparison because it has no controlled intensity sweep.


\subsection{Weather Severity Is Not a Proxy for Perception Difficulty}
\label{sec:intensity_analysis}

Commanded weather intensity is a reliable control variable for visual corruption, but not for robotic perception difficulty. Within each weather type, increasing intensity produces an almost monotonic increase in RGB degradation (Fig.~\ref{fig:intensity_analysis}, left). Yet frozen detection performance shows only weak and often non-monotonic dependence on the same intervention (Fig.~\ref{fig:intensity_analysis}).

This decoupling shows that physical corruption and robotic perception difficulty are distinct. The latter depends jointly on degradation type, scene content, restoration behavior, and downstream model response. Weather intensity should therefore be treated as a controlled intervention, not a universal difficulty scale, motivating direct task-level evaluation of perception robustness under extreme conditions.

\section{Conclusion}\label{sec_conclusion}

This work studies how complementary onboard sensing can preserve robotic perception when extreme environmental conditions severely degrade the primary RGB observation. We introduce SenseWeather, a controlled multimodal benchmark spanning diverse operational environments, extreme weather and illumination conditions, and grounded high-level perception tasks. We further develop Fusion-to-RGB through two parameter-efficient sensor-fusion designs that recover a conventional RGB observation for unchanged pretrained perception models. Across open-vocabulary grounding and vision-language reasoning, multimodal sensing consistently improves downstream perception over RGB-only restoration. Our analysis further shows that neither conventional image fidelity nor nominal weather severity reliably predicts perception robustness, highlighting the importance of evaluating extreme-condition robotic perception directly at the task level.

\section*{Acknowledgment}
This work was supported by the ONR Global X-Challenge Grant (N62909-25-1-2067).

{
    \small
    \bibliographystyle{IEEEtran}
    \bibliography{main}
}

\end{document}